\documentclass[lettersize,journal]{IEEEtran}
\usepackage{amsmath,amsfonts}
\usepackage{algorithmic}
\usepackage{algorithm}
\usepackage{array}
\usepackage[caption=false,font=normalsize,labelfont=sf,textfont=sf]{subfig}
\usepackage{textcomp}
\usepackage{stfloats}
\usepackage{url}
\usepackage{verbatim}
\usepackage{graphicx}
\usepackage{cite}
\usepackage{multirow}
\usepackage{bm}
\usepackage{booktabs} 
\usepackage{caption}
\usepackage{subfig}

\newcommand{\itm}[1]{\textrm{\textit{#1}}}  
\newcommand{\angletoken}[1]{$\langle$\texttt{#1}$\rangle$}  
\newcommand{\squaretoken}[1]{\texttt{[#1]}}  

\begin{document}

\title{Semantic-Enhanced Explainable Finetuning for Open-Domain Dialogues}

\author{Yinhe Zheng, Yida Wang, Pei Ke, Zhenyu Yang, Minlie Huang

\thanks{zhengyinhe1@163.com}
\thanks{kepei1106@outlook.com}
\thanks{aihuang@tsinghua.edu.cn}

}

\markboth{IEEE/ACM TRANSACTIONS ON AUDIO, SPEECH, AND LANGUAGE PROCESSING}%
{Shell \MakeLowercase{\textit{et al.}}: A Sample Article Using IEEEtran.cls for IEEE Journals}


\maketitle

\begin{abstract}

This paper propose to combine pretrained language models with the modular dialogue paradigm for open-domain dialogue modeling. Our method, \textit{semantic-enhanced finetuning}, instantiates conversation \textit{understanding}, \textit{planning}, and \textit{response generation} as a language model finetuning task. At inference, we \textit{disentangle} semantic and token variations by specifying sampling methods and constraints for each module separately. For training and evaluation, we present X-\textsc{Weibo}, a Chinese multi-turn open-domain dialogue dataset with \textit{automatic} annotation for emotions, DAs, and topical words. Experiments show that semantic-enhanced finetuning outperforms strong baselines on non-semantic and semantic metrics, improves the human-evaluated relevance, coherence, and informativeness, and exhibits considerable controllability over semantic variables.
\footnote{Under review, code and data will be publicly available.}
\end{abstract}


\section{Introduction}
\label{sec:introduction}

\IEEEPARstart{B}{uilding} open-domain conversational agents is a long-standing goal of artificial intelligence \cite{HuangZG20}. Recently, open-domain dialogue systems built upon large-scale generative pretrained language models \cite{ZhangSGCBGGLD20,Roller20Blender,Adiwardana20Meena,bao2021plato} achieve state-of-the-art conversation performances in terms of the quality, relevance, engagingness, and diversity. In the development of conversational agents, what has long been studied is the \textit{modular} dialogue modeling \cite{roy2000spoken,tur2011spoken}, which explicitly models the \textit{understanding} and \textit{planning} of \textit{semantic} information \cite{schank1977scripts,carberry1990plan} that flows throughout the conversation. This modular idea has been explored in task-oriented dialogue systems \cite{WilliamsRH16a,li2017end,rastogi2018multi,Peng20SOLOIST}, knowledge-driven conversations \cite{SuSZZCZNZ20,hedayatnia2020policy}, negotiation \cite{HeCBL18}, and persuasion \cite{SanthanamCMDBHZ20}. A benefit of this paradigm is its explainability and controllability \cite{komeili2021internet,thoppilan2022lamda}. It also allows one to specify the type of modeled semantic variable, which makes the system robust to other variations.

\begin{figure*}[t]
	\centering
	\includegraphics[width=0.7\linewidth]{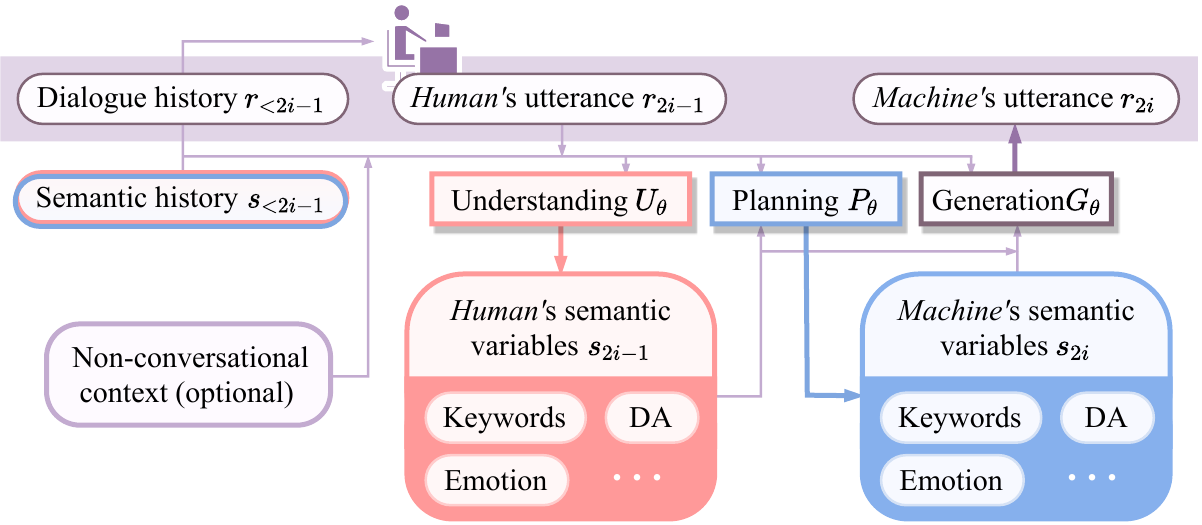}
	\caption{\label{fig:semantics-dialogue} Modular dialogue modeling. In each turn, the model reads \textit{Human}'s last utterance $r_{2i-1}$, predicts the semantic variables $s_{2i-1}$ of this utterance (\textit{understanding} $U_{\theta}$), predicts the semantic variables $s_{2i}$ of its response (\textit{planning} $P_{\theta}$), and generates the response $r_{2i}$ (\textit{response generation} $G_{\theta}$). }
\end{figure*}

In this paper, we propose to bridge the pretrained language models with the modular dialogue modeling paradigm for \textit{open-domain} dialogue systems, which we name as \textit{semantic-enhanced finetuning}. We model the understanding and planning of \textit{emotions}, \textit{dialogue acts} (DAs), and \textit{topics} during the finetuning of open-domain dialogue models. We use a pretrained language model to instantiate three modules: \textit{understanding}, \textit{planning}, and \textit{response generation}. Specifically, the \textit{understanding} module infers the semantic variables of the dialogue history, the \textit{planning} module plans the semantic variables of the next response, and the \textit{response generation} module generates diverse responses. An overview is shown in Figure \ref{fig:semantics-dialogue}. These three modules share the pretrained parameters and are jointly finetuned using the cross-entropy loss. At inference, we \textit{disentangle} semantic and token variations by specifying sampling methods and constraints for each module separately. Specifically, we find that the \textit{minimal length constraint} is effective for topic planning and response generation. We also design a \textit{repetition constraint} for topic planning to address the repetition problem. 

Semantic-enhanced finetuning has three benefits.
\textbf{First}, it exploits the response generation capability acquired during pretraining, which improves the response generation step of the modular paradigm.
\textbf{Second}, it equips chatbots with fine-grained \textit{controllability}: One may manipulate the \textit{plans} to control the emotions, DAs, and topics of the response; One may also keep the \textit{plans} unchanged and manipulate the \textit{response generation} module to generate stylized responses while preserving the emotions, DAs, and topics.
\textbf{Third}, semantic-enhanced finetuning is \textit{explainable}. For example, when an improper response is generated, one can \textit{attribute the error} by checking the predicted semantic variables.

Our model should be finetuned on conversations annotated with semantic variables, but human annotation for large-scale conversation data faces the \textit{scalability} challenge. Thus, we propose collecting a set of manually annotated dialogues and then automatically annotating semantic variables with pretrained classifiers. Based on conversations collected from Weibo, we present X-\textsc{Weibo}, a Chinese multi-turn open-domain dialogue dataset with over 500k dialogue sessions, annotated with \textit{emotions}, \textit{dialogue acts}, and \textit{topical words}. 

We compare our method with strong baselines using the same pretrained weights and observe that it has the best BLEU scores \cite{PapineniRWZ02}, embedding-based metrics \cite{LiuLSNCP16}, diversity \cite{LiGBGD16}, and semantic-level agreement with human references. Human evaluation results show that our approach significantly outperforms the baselines regarding relevance, coherence, and informativeness. For controllability, we show that our method can guide the generation of responses by controlling the semantic variables. Our contributions are as follows:
\begin{itemize}
    \item We propose semantic-enhanced finetuning for open-domain dialogue systems, which bridges large-scale pretrained models with modular dialogue modeling. 
    \item We present the X-\textsc{Weibo} dataset with automatic annotation for emotions, DAs, and topical words, which will be open-sourced.
    \item In the experiments, our method outperforms strong baselines in automatic metrics and human evaluation.
\end{itemize}

\section{Related Work}
\label{sec:related-work}

Modeling semantic variables in conversation has a long history \cite{schank1977scripts,carberry1990plan}. Prior works in this direction build modular systems to capture dialogue states \cite{roy2000spoken,li2017end}, and various approaches focus on the task of understanding and planning in open-domain dialogues \cite{tur2011spoken,rastogi2018multi}. However, most of these systems build separated models for the understanding, planning, and generation modules, while we instantiate these modules as a unified pre-trained language model. 

Transformers \cite{VaswaniSPUJGKP17} pretrained on large-scale conversation data achieve state-of-the-art performances on the open-domain dialogue modeling \cite{golovanov2019large,wolf2019transfertransfo}, and examples include DialoGPT \cite{ZhangSGCBGGLD20}, Blender \cite{Roller20Blender,komeili2021internet}, Meena \cite{Adiwardana20Meena}. These models view conversation data as sequences of tokens alternating between speakers and cast open-domain dialogue modeling as a language modeling task. Such simplicity enables scalability to large-scale conversation data. However, to truly understand human conversations, an agent should be able to connect utterances to explainable semantic concepts. Our focus is to combine the modular dialogue paradigm, which models conversational understanding and planning, with pretrained language models for open-domain dialogue systems.

It also worth noting that some works have studied the planning of general dialogue acts (DAs) \cite{Xu18DialogueActs}, and others have designed DAs for specific domains, e.g., dialogue states \cite{WilliamsRH16a,Peng20SOLOIST,budzianowski2019hello,wu2020tod} for task-oriented dialogues, strategies \cite{HeCBL18} for negotiation, lexical-conceptual structures \cite{SanthanamCMDBHZ20} for persuasion, and policy or intent planning \cite{SuSZZCZNZ20,hedayatnia2020policy} that for knowledge-driven conversations. A recent work \cite{ghazarian2021discol} extracts entities and topics and plans keywords on the knowledge-grounded TopicalChat dataset \cite{GopalakrishnanH19}. However, a prerequisite for these approaches is that the dialogue acts modeled in these works are specifically designed for different domains. In this study, we focus on general semantic variables shared by \textit{most} conversations: emotions \cite{Zhou2018EmotionalCM,varshney2021modelling}, general dialogue acts \cite{LiSSLCN17}, and topical words \cite{LiuCZS11}, which are more general and more transferable in the open domain.

Our work is also related to CVAE \cite{ZhaoZE17} and PLATO \cite{BaoHWWW20,Bao20PLATO2,bao2021plato}, which captures the discourse-level diversity in human conversations with latent variables. Compared with them, the modular paradigm has a higher level of controllability (e.g., it can be intervened with human-specified plans) and explainability (e.g., it can be debugged by checking the predicted plans). Meanwhile, our semantic variables (i.e., emotions, DAs, and topical words) are more interpretable than Gaussian or categorical latent variables.

\section{Our Approach}
\label{sec:method}

\subsection{Task Formulation}
\label{subsec:formulation}
A multi-turn dialogue session is a sequence of utterances alternating between \textit{Human} and \textit{Machine}: $(r_{1}, \ldots, r_{2N})$. Without loss of generality, our formulation assumes that \textit{Human} initiates the conversation, i.e., $r_{2i-1}$ and $r_{2i}$ ($i = 1, \ldots, N$) are uttered by \textit{Human} and \textit{Machine}, respectively. Each session may also be grounded on some non-conversational context, e.g., persona and topics. 

\noindent\textbf{Semantic variables \ \ } Each \textit{Human}'s or \textit{Machine}'s utterance $r_{i}$ ($i = 1, \ldots, 2N$) can be explained by $K$ semantic variables $s_{i} = \{\langle\mathrm{s\_key}^{k}, \mathrm{s\_val}^{k}_{i}\rangle\}_{k=1}^{K}$, where $\mathrm{s\_key}^{k}$ is the type, and $\mathrm{s\_val}^{k}_{i}$ is the value. In our X-\textsc{Weibo} dataset introduced in Section~\ref{sec:data}, $K=3$ types of semantic variables are annotated for each utterance: topical words, DAs, and emotions. Specifically, the value of topical words is a list of deduplicated phrases (each phrase is a list of tokens); the value of DAs (emotions) is a list of DA (emotion) labels, each label corresponding to a sentence in the utterance. For example, if an utterance first answers a question and then post another question, then its DAs value is $[$\textit{Inform}$,$ \textit{Question}$]$. 

\noindent\textbf{Dialogue decomposition \ \ } With the semantic variables, modular dialogue modeling decomposes open-domain dialogues into three modules: \textit{understanding} $U_{\theta}$, \textit{planning} $P_{\theta}$, and \textit{response generation} $G_{\theta}$. Specifically, the \textit{understanding} module infers the semantic variables of \textit{Human}'s last utterance. The \textit{planning} module plans the semantic variables of \textit{Machine}'s response. The \textit{response generation} module generates responses, which captures the syntactic variations that are not covered by the plan. Formally, the overall objective to be maximized is 
\begin{equation}
\label{eq:formulation-objective}
\begin{split}
    \mathbb{E}_{\mathcal{D}}\Big[\sum_{i=1}^{N}&\Big(\log U_{\theta}(s_{2i-1}|r_{\leq 2i-1}, g, s_{<2i-1}) \\
    & + \log P_{\theta}(s_{2i}|r_{\leq 2i-1}, g, s_{\leq2i-1}) \\
    & + \log G_{\theta}(r_{2i}|r_{\leq 2i-1}, g, s_{\leq2i}) \Big)\Big]
\end{split}
\end{equation}
where $g$ is the non-conversational contexts and $\mathcal{D}$ is the empirical distribution of data, e.g., the X-\textsc{Weibo} dataset introduced in Section~\ref{sec:data}. 

\subsection{Semantic-Enhanced Finetuning}
\label{subsec:model}
We propose to model the above three modules with the same generative pretrained weights, which we name as \textit{semantic-enhanced finetuning}. Based on Eq.~(\ref{eq:formulation-objective}), we place the non-conversational context, utterances, and semantic variables in the order:
\begin{equation}
\label{eq:order-of-sequences}
    g, \underbrace{r_{1}, s_{1}, s_{2}, r_{2}}_{\textrm{the 1$^{\itm{st}}$ turn}} \ldots \underbrace{r_{2N-1}, s_{2N-1}, s_{2N}, r_{2N}}_{\textrm{the $N^{\itm{th}}$ turn}}.
\end{equation}
The non-conversational context $g$ is placed at the beginning since they are shared by all \textit{Machine}'s utterances. In the $i^{\itm{th}}$ turn, the model infers the semantic variables $s_{2i-1}$\footnote{Note that $s_{2i-1}$ contains key-value pairs for several variables, e.g., DAs, emotions, and topical words.} of \textit{Human}'s last utterance $r_{2i-1}$, (\textit{understanding}), predicts the semantic variables $s_{2i}$ of \textit{Machine}'s next utterance (\textit{planning}), and generates the next utterance $r_{2i}$ (\textit{response generation}). We define five token types \cite{DevlinCLT19} to distinguish the elements in Eq.(\ref{eq:order-of-sequences}), which include \textit{Human}'s utterances, \textit{Machine}'s utterances, \textit{Human}'s semantic variables, \textit{Machine}'s semantic variables, and non-conversational context. Based on Eq.~(\ref{eq:order-of-sequences}), the three modules $U_{\theta}$, $P_{\theta}$, and $G_{\theta}$ are unified within a single sequence generation model (e.g., a generative LM or a seq2seq model) that can be trained end-to-end. When predicting each element in Eq.~(\ref{eq:order-of-sequences}), the model is conditioned on all elements in the prefix. By switching the roles of \textit{Human} and \textit{Machine}, we derive \textit{two} samples in the form of Eq.~(\ref{eq:order-of-sequences}) from each dialogue session. 

\begin{table}[t]
\centering
\small
\begin{tabular}{@{}l|c@{}}
\toprule
\multirow{2}{*}{Semantic variable keys} & \angletoken{topical} \ \ \angletoken{emotion} \\
& \angletoken{dialog\_act}  \\
\midrule
List separator & \angletoken{list\_sep} \\
End of key-value pair & \angletoken{eokv} \\
\midrule
Start of conversation & \squaretoken{CLS} \\
Start of \textit{Human}'s utt. & \angletoken{human} \\
Start of \textit{Machine}'s utt. & \angletoken{machine} \\
End of utterance & \squaretoken{SEP} \\
\bottomrule
\end{tabular}
\caption{\label{tab:special-tokens} Special tokens used in the model}
\end{table}

\begin{figure*}[t]
	\centering
	\includegraphics[width=\linewidth]{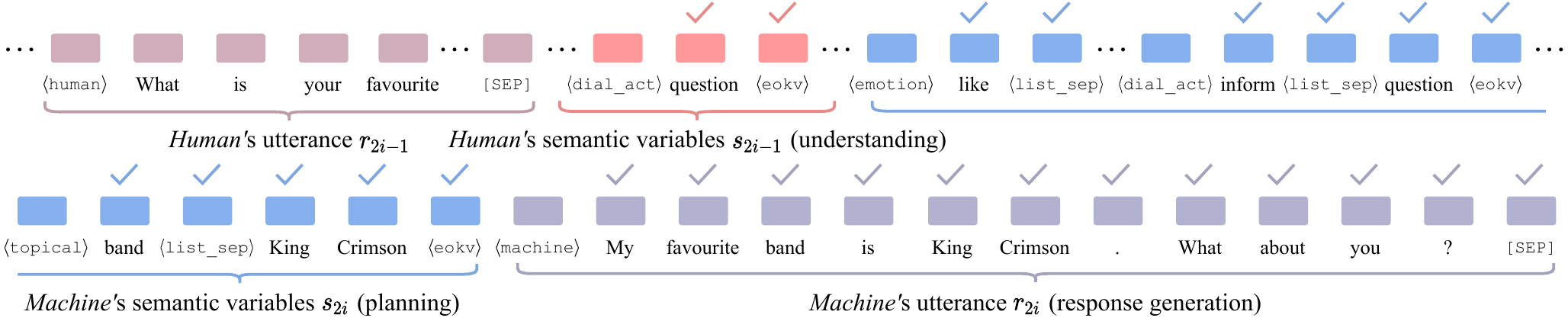}
	\caption{\label{fig:input-scheme} The scheme we used to convert Eq.~(\ref{eq:order-of-sequences}) into tokens. The type of sequences and the module that learn them are under the brackets. When predicting each sequence, all preceding tokens are used as the condition. We compute loss for tokens with a tick above. Denotations of special tokens are shown in Table~\ref{tab:special-tokens}.}
\end{figure*}

It still remains an open question how to represent \textit{structured} semantic variables as a sequence. In this paper, we design several special tokens to linearize the structured variables. The special tokens are shown in Table~\ref{tab:special-tokens}, and a linearized input is shown in Figure~\ref{fig:input-scheme}. We assign a special token for each $\mathrm{s\_key}^{k}$ and use \angletoken{eokv} to mark the end of a key-value pair. Since the values $\mathrm{s\_val}^{k}_{i}$ are all lists (Section~\ref{sec:data}), we define a list separator token \angletoken{list\_sep} to separate items in a list. In Figure~\ref{fig:input-scheme}, for example, \textit{Machine}'s utterance ``\textit{My favorite band is KC. What about you?}'' has two DAs: \textit{Inform} and \textit{Question}; thus, the DAs-value to be planned is linearized as ``\textit{Inform} \angletoken{list\_sep} \textit{Question}''. We place a \squaretoken{CLS} between the non-conversational context and \textit{Human}'s first utterance. We use \angletoken{human} or \angletoken{machine} to denote the speaker, and \squaretoken{SEP} stands for the end of an utterance. We place the semantic variables in a predefined order: emotion, dialogue act, and topical words, but note that different orders factorize the \textit{same} joint distribution over the semantic variables. 

To optimize the objective function in Eq.~(\ref{eq:formulation-objective}), we compute the autoregressive cross-entropy loss for the linearized semantic variables and \textit{Machine}'s utterances in Eq.~(\ref{eq:order-of-sequences}). Additionally, we do not compute loss for the semantic variables' keys $\mathrm{s\_key}^{k}$ since they do not need to be predicted during inference. In the example in Figure~\ref{fig:input-scheme}, we put a tick over the tokens that we compute loss for. Our approach can adopt any sequence generation model.

\subsection{Inference}
At inference, tokens that we do not compute loss for are encoded or used as prompts for decoding. In each turn, three sequences need to be decoded. The \textit{understanding} module first infers the semantic variables of \textit{Human}'s last utterance, the \textit{planning} module then plans the semantic variables for the next utterance, and finally the \textit{response generation} module generates the response. We \textit{disentangle} semantic-level and token-level variations by specifying \textit{sampling methods} and \textit{constraints} for each module separately, which is detailed as follows. 

\subsubsection*{\textit{Understanding} Decoding} 
For the \textit{understanding} module, we do not set the minimum lengths, and we set the lengths for the linearized topical words, emotions, and DAs as 20, 10, 10. Greedy decoding is adopted. 

\subsubsection*{\textit{Planning} Decoding} 
For the \textit{planning} module, we set the minimum (maximum) lengths for the linearized topical words, emotions, and DAs as 5 (20), 0 (10), 0 (10), and greedy decoding is used. The non-zero minimum length for topical words enforces non-trivial topical words to be generated, which is expected to improve the informativeness of the response. 

\noindent\textbf{Repetition constraint for topical \textit{planning} \ } We introduce the \textit{repetition constraint} to avoid repeated topical words to be generated. Specifically, we suppress an $n$-gram prefix of a topical word to be generated if this prefix has been generated in the current plan. Note that the \textit{repetition constraint} is not used for response generation since directly suppressing repeated $n$-grams in the response unavoidably causes disfluency, e.g., a grammatical sentence may contain two ``\textit{of the}''. Since our topical words only contain deduplicated informative words, we can more safely suppress repeated $n$-grams in topical words without sacrificing fluency. 

\subsubsection*{\textit{Response Generation} Decoding} 
For \textit{response generation}, we use top-$k$ sampling \cite{LewisDF18} and top-$p$ sampling \cite{HoltzmanBDFC20} with temperature $\tau$. We set $k=50$, $p=0.9$, and $\tau=0.7$, which are shared by all models and baselines in Section~\ref{sec:experiments}. In our implementation, the length constraints are achieved by manipulating the predicted probability of \angletoken{eokv} (for \textit{understanding} and \textit{planning}) or \squaretoken{SEP} (for \textit{response prediction}) as $0$ or $1$. 

\section{X-\textsc{Weibo} Dataset}
\label{sec:data}

\subsection{Conversation Collection}

Conversation data annotated with semantic variables is a prerequisite for our method. In this section, we introduce the X-\textsc{Weibo} dataset that is collected to facilitate our study. Dialogues in X-\textsc{Weibo} consist of comments issued by Weibo\footnote{Weibo is one of the largest Chinese social platforms.} users. Each comment is regarded as an utterance in a dialogue, and the reply relations between these comments are used to construct dialogues in X-\textsc{Weibo}. We use the data processing and cleaning pipelines proposed by \cite{WangKZHJZH20} to filter low-quality dialogues. After filtering, dialogues are split into the training, validation, and test splits. Dataset statistics are shown in \ref{tab:dataset-stats}. 

\begin{table*}[t]
\centering
\small
\begin{tabular}{@{}lccccc@{}}
\toprule
 & Sessions & Utt./Session & Tokens/Utt. & DAs(Emotions)/Utt. & Topical Words/Utt. \\
\midrule
Train  &  500K & 5.20 & 15.52 & 1.23 & 1.14 \\
Valid  &  20K  & 5.19 & 15.55 & 1.23 & 1.15 \\
Test   &  10K  & 5.21 & 15.74 & 1.24 & 1.16 \\
\bottomrule
\end{tabular}
\caption{\label{tab:dataset-stats} Dataset statistics. DA and Utt. stand for \textit{dialogue act} and \textit{utterance}, respectively.}
\end{table*}

\subsection{Semantic Variables Annotation}

The optimal way to acquire annotations for the semantic variables is to recruit well-trained human annotators with profound knowledge about linguistics. However, human annotation faces the \textit{scalability} challenge for large-scale conversation data (e.g., over 3.3M sentences in X-\textsc{Weibo}). Thus, we propose to annotate semantic variables with pretrained classifiers automatically.
Specifically, three kinds of semantic variables are annotated: topical words, dialogue acts (DAs), and emotions since they carry important semantic information in human conversations. The annotation process for each variable is detailed as follows. 

\noindent\textbf{Topical words \ \ } The topical words for each utterance in X-\textsc{Weibo} are extracted using the THUCKE package \cite{LiuCS11}, which uses a word trigger method to extract topical words with the help of a learned word alignment table. We first concatenate all utterances in X-\textsc{Weibo} to construct the input file for THUCKE and then use THUCKE to extract topical words for the entire corpus. We keep the most frequent 6,000 topical words as the final topical word vocabulary and align them to each utterance. Specifically, the topical words for an utterance are words that appear both in this utterance and the extracted topical word vocabulary.

\noindent\textbf{Sentence split \ \ } In this study, the labels for DA and emotion are annotated in a \textit{finer-grained sentence-level} rather than the utterance-level. Specifically, we first split each utterance (i.e., each turn) in a dialogue session into several sentences based on the end-of-sentence punctuation marks such as the period or question marks. Then we obtain the DA and emotion label for each split sentence using pre-trained DA and emotion classifiers. This annotation scheme is designed based on our observations for dialogues in X-\textsc{Weibo}, i.e., each dialogue utterance in X-\textsc{Weibo} may correspond to several different DA or emotion labels. The sentence-level annotations of DA and emotion enable us to capture finer-grained semantics in our dialogue model.

\noindent\textbf{Dialogue acts \ \ }  To obtain DA labels, we adopt the DA scheme employed in the DailyDialog dataset \cite{LiSSLCN17}. Specifically, four categories of DA are considered: \textit{Inform}, \textit{Question}, \textit{Directive}, and \textit{Commissive}. This study first extracts sentences with DA annotations from the DailyDialog dataset, which yields about 102.9K annotated sentences. Then we divide these sentences into the train and test set with the size of 101.0K and 1.9K, respectively. A DA classifier is built by fine-tuning the BERT-base \cite{DevlinCLT19} model on the training set. The resulting classifier achieves an accuracy score of 84.58\% on the testing set, and it is further used to predict the DA labels for each sentence in our X-\textsc{Weibo} dataset.

\noindent\textbf{Emotions \ \ }  Emotions are annotated similar to DAs. Specifically, eight emotion categories are considered in our study: Fear, Surprise, Anger, Disgust, Like, Happiness, Sadness, and None, where the label None corresponds to sentences that do not carry obvious emotions. For training, we merge two publicly available datasets\footnote{http://tcci.ccf.org.cn/conference/2013/}$^,$\footnote{https://github.com/MingleiLI/emotion\_corpus\_weibo} \cite{li2016emotion} with such annotations and extract 83.29K annotated sentences. We split these data into the train and test set with the size of 80.0K and 3.29K, respectively, and obtain an emotion classifier by fine-tuning the BERT-base model. The resulting classifier achieves an accuracy score of 63.71\% on the testing set.

Note that the automatic annotation scheme proposed above is a preliminary attempt towards scalable semantic annotations. We leave more advanced annotation methods for future studies (discussed in Section~\ref{sec:discussion}).

\begin{figure}[t]
	\centering
	{
        \centering
        \includegraphics[width=0.85\linewidth]{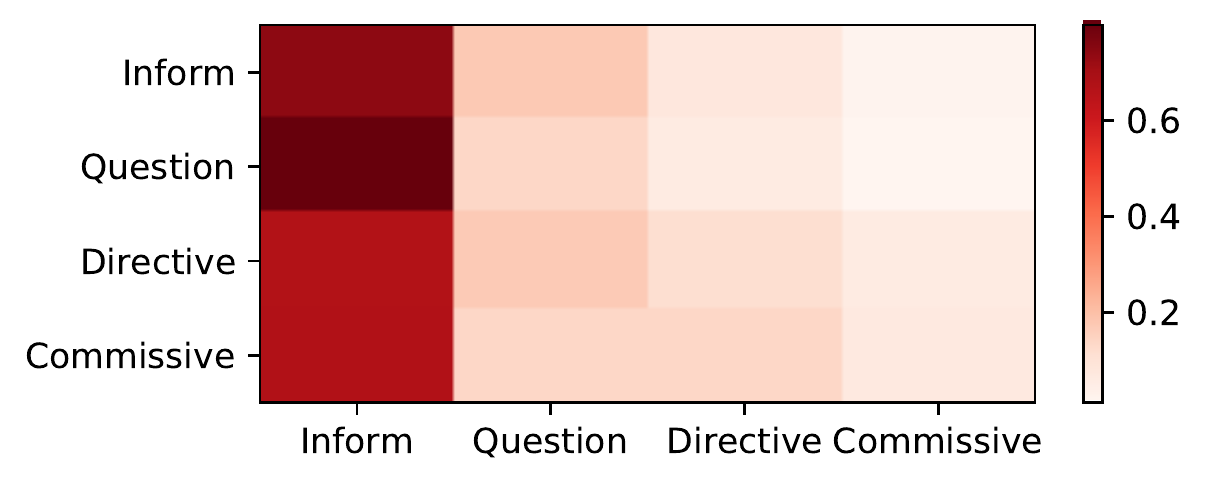}
		}
	
	{
        \centering
        \includegraphics[width=0.8\linewidth]{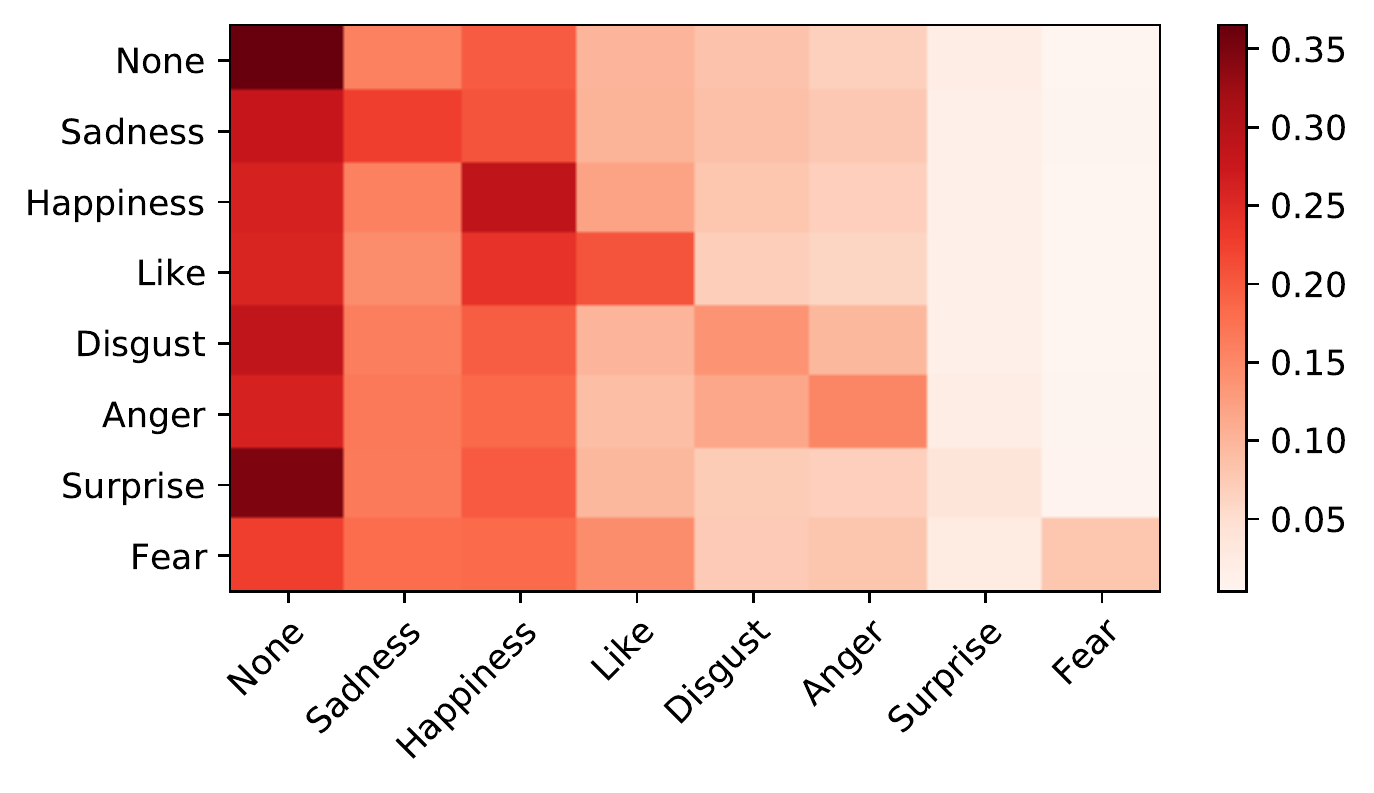}
		}
	\caption{\label{fig:transition-matrices} DA (upper) and emotion (lower) label transitions. $y$-axis and $x$-axis represent the previous and the current utterance, respectively (e.g., the first row in the upper part shows the distribution of the DA of the current utterance when the DA of the previous utterance is \textit{Inform}). For readability, we use utterance-level DAs and emotions in this figure, without the sentence split introduced in Section~\ref{sec:data}.}
\end{figure}

\noindent\textbf{Data verification \ \ } To validate the quality of the automatic annotation for DA and emotion, we conduct human evaluations on 1000 randomly sampled sentences from X-\textsc{Weibo}. Each sentence is annotated by three independent annotators, and the majority vote determines the final annotation. The results indicate that 77.5\% automatic emotion labels and 88.7\% automatic DA labels are verified by human annotators. This proves that the automatic annotations for DAs and emotions in X-\textsc{Weibo} are of relatively high quality. We measure the agreement between each annotator using free-marginal kappa \cite{fleiss1971measuring}, and the kappa score for the emotion and DA annotation is 0.49 (moderate agreement) and 0.75 (substantial agreement), respectively. We assume that the gap between the inter-annotator agreements of emotion and DA annotation can be attributed to the fact that emotions are more ambiguous for humans to distinguish. We also visualize the transition patterns of the DA and emotion labels in Figure~\ref{fig:transition-matrices}.

\begin{table*}[t]
\centering
\small
\begin{tabular}{@{}l@{}cccccccc@{}}
\toprule
~  & \multicolumn{4}{c}{Token-based} & \multicolumn{2}{c}{Embedding-based} & \multicolumn{2}{c}{Diversity} \\
\cmidrule(r){2-5} \cmidrule(lr){6-7} \cmidrule(l){8-9}
~  			                                & BLEU-1 & BLEU-2 & BLEU-3 & PPL & Average & Extreme & Dist-1 \% & Dist-2 \% \\
\midrule
GPT-2                                       & \underline{11.5}  & \underline{5.0}   & \underline{2.6}   & 21.2  & \underline{0.788} & \underline{0.747} & \bf 0.61          & 10.42 \\
GPT-2 + \textit{min length}                 & 12.7              & 5.5               & 2.8               & 21.2  & \bf 0.807         & 0.779             & \underline{0.48}  & \underline{8.86} \\
SE-GPT-2 (ours)                            & \bf 13.2          & \bf 6.2           & \bf 3.1           & ---   & \bf 0.806         & \bf 0.785         & 0.55              & \bf 11.57 \\
\quad w/ \textit{gold variables}$^{[1]}$    & 29.5              & 19.6              & 12.1              & 12.5  & 0.842             & 0.827             & 0.58              & 12.23 \\
\midrule
CDialGPT-2                                  & \underline{10.9}  & \underline{4.8}   & \underline{2.4}   & 23.5  & \underline{0.784} & \underline{0.742} & \bf 0.54          & 9.62 \\
CDialGPT-2 + \textit{min length}            & 12.4              & 5.3               & 2.6               & 23.5  & \bf 0.805         & 0.775             & \underline{0.41}  & \underline{7.96} \\
SE-CDialGPT-2 (ours)                       & \bf 13.1          & \bf 6.1           & \bf 3.1           & ---   & \bf 0.806         & \bf 0.783         & 0.47              & \bf 10.53 \\
\quad w/ \textit{gold variables}$^{[1]}$    & 29.0              & 19.3              & 12.0              & 13.5  & 0.842             & 0.826             & 0.49              & 11.38 \\
\bottomrule
\end{tabular}
\caption{\label{tab:word-level-automatic-results} Non-semantic automatic evaluation. $^{[1]}$Using gold semantic variables for the controllability test. For models that use planned semantic variables, best results are shown in \textbf{bold}, and worst results are \underline{underlined}.}
\end{table*}

\section{Experiments}
\label{sec:experiments}

\subsection{Experimental Setting}
We conduct experiments using \textit{pairwise comparison}. We use two pretrained LMs as the backbones: 1) Chinese GPT-2 \cite{Radford2019GPT2}\footnote{huggingface.co/uer/gpt2-chinese-cluecorpussmall}, which is pretrained on CLUECorpusSmall \cite{Xu2020CLUECorpu2020}, and 2) CDialGPT-2\footnote{huggingface.co/thu-coai/CDial-GPT2\_LCCC-base}, which is pretrained on a large-scale Chinese conversation dataset named LCCC-base \cite{WangKZHJZH20}. \textbf{We make sure that \textit{none} of the pretraining data overlap with X-\textsc{Weibo}}. For each backbone, the vanilla model (GPT-2 and CDialGPT-2) is optimized with the language modeling loss without using the semantic variables, as done in DialoGPT \cite{ZhangSGCBGGLD20}. We observe that setting a minimal decoding length improves most automatic metrics, which echoes the observation by \cite{Roller20Blender}. Thus, we also include a \textit{min length} version of each backbone (GPT-2 + \textit{min length} and CDialGPT-2 + \textit{min length}) as a baseline. For a fair comparison, hyperparameters are shared by all models whenever possible. The minimal decoding length is set as 9, and the maximum decoding length is set as 32, which are tuned on the validation data. 

Note that the data distribution of X-\textsc{Weibo} is different from the pretraining data, i.e., CLUECorpusSmall \cite{Xu2020CLUECorpu2020} and LCCC-base \cite{WangKZHJZH20}. Specifically, CLUECorpusSmall is used for general pretraining, and most sessions in LCCC-base are single-turn. 

Models are validated for every 5000 steps, based on the perplexity (PPL) on the validation set. The batch size is 24. We use Adam \cite{KingmaB14} with the initial learning rate $5\times 10^{-5}$ and gradient clip with the norm as $1.0$. The learning rate decays by half when the validation PPL does not improve five times, and training terminates after three decays. Each experiment is run on four TITAN X (Pascal) GPUs or four TITAN Xp GPUs. All models take around two days to converge. 

\subsection{Automatic Evaluation of Response Generation}
\label{subsec:automatic-evaluation}

We report several automatic metrics widely used by previous works, BLEU-\{1,2,3\} \cite{PapineniRWZ02}, and Embedding-\{Average, Extreme\} \cite{LiuLSNCP16}, which show the word-level similarity between the predictions and human references. To evaluate the diversity of model predictions, we adopt the Dist-\{1,2\} metrics \cite{LiGBGD16}. We do not report the perplexity (PPL) for our methods since it requires an intractable marginalization over all possible semantic plans. The embedding-based metrics use the Jieba parser\footnote{https://github.com/fxsjy/jieba} to parse the responses into Chinese phrases (for accurate embedding lookup), and all other metrics view each Chinese character as a word. For the embedding-based metrics, we use the Chinese phrase embeddings by \cite{SongSLZ18}
for embedding lookup. 

Table~\ref{tab:word-level-automatic-results} shows similar patterns for the two backbones. Specifically, the vanilla models (GPT-2 and CDialGPT-2) have the worst BLEU scores and the worst embedding-based scores. The performances of the \textit{min length} models show that by enforcing a minimum length for the responses, we can improve the BLEU scores and embedding-based scores, but the diversity is greatly sacrificed. Our SE-GPT-2 and SE-CDialGPT-2 models largely improve the BLEU scores and embedding-based scores while maintaining comparable diversity with the vanilla models. This result indicates that responses generated by our models are more informative than the \textit{min length} models. Given the fact that the only difference between our models and the \textit{min length} models is the use of semantic variables, we claim that our method enables \textit{long} responses to be \textit{informative} as well. Additionally, results suggest that uni-gram diversity and bi-gram diversity are not correlated. It indicates that diversity is not only to generate rare words but also to generate diverse compositions of words (e.g., bi-grams). 

\begin{table*}[t]
\centering
\small
\begin{tabular}{@{}lccc@{}}
\toprule
~  & Topical-Recall  & DAs-F1  & Emotions-F1 \\
\midrule
GPT-2                                       & \underline{4.1}   & 47.1              & 19.8 \\ 
GPT-2 + \textit{min length}                 & 4.9               & \underline{45.5}  & \underline{19.0} \\
SE-GPT-2 (ours)                              & \bf 13.1          & \bf 52.1          & \bf 22.2 \\ 
\quad w/ \textit{gold variables}$^{[1]}$    & 91.8              & 89.7              & 69.8 \\ 
\midrule
CDialGPT-2                                  & \underline{3.7}   & 47.0              & 19.9 \\
CDialGPT-2 + \textit{min length}            & 4.3               & \underline{46.0}  & \underline{19.1} \\
SE-CDialGPT-2 (ours)                         & \bf 12.7          & \bf 52.1          & \bf 22.4 \\
\quad w/ \textit{gold variables}$^{[1]}$    & 92.5              & 89.2              & 67.6 \\
\bottomrule
\end{tabular}
\caption{\label{tab:semantic-level-automatic-results} Semantic-level automatic evaluation. $^{[1]}$Using gold semantic variables for the controllability test. For models that use planned semantic variables, best results are shown in \textbf{bold}, and worst results are \underline{underlined}.}
\end{table*}

\begin{table*}[t]
\centering
\small
\begin{tabular}{@{}lccc@{}}
\toprule
~  & Relevance and coherence  & Informativeness & Engagingness \\
\midrule
Human                           & 2.27                          & 2.05                          & 1.98 \\ 
\midrule
GPT-2                           & \underline{2.08}              & \underline{1.84}              & \underline{1.81} \\ 
GPT-2 + \textit{min length}     & 2.13                          & 1.95                          & 1.93 \\ 
SE-GPT-2 (ours)                  & \ \ \ \bf 2.20$^{\dag\ddag}$  & \ \ \ \bf 2.07$^{\dag\ddag}$  & \ \ \bf 1.94$^{\dag}$ \\
\bottomrule
\end{tabular}
\caption{\label{tab:human-results} Human evaluation results. The free-marginal kappa score for relevance and coherence, informativeness, and engagingness is 0.79, 0.70, and 0.74, respectively, which all show substantial agreement. $^{\dag}$ and $^{\ddag}$ denotes that our model is significantly better than GPT-2 and GPT-2 + \textit{min length}, respectively ($p<0.01$ with paired two-sample \textit{t}-test). Best results are shown in \textbf{bold}, and worst results are \underline{underlined}.}
\end{table*}

Besides the conventional evaluation above, we also evaluate the generated responses at the \textit{semantic level}. For topical words, we report the \textit{\textbf{Topical-Recall}}, which is the proportion of labeled topical words in X-\textsc{Weibo} that appear in the generated response. For the DAs and emotions, we first split the generated response into sentences based on the protocol in Section~\ref{sec:data}, which allows us to deal with utterances that contain more than one sentence with different DAs or emotions. We use the pretrained DA / emotion classifiers in Section~\ref{sec:data} to classify each sentence. As mentioned in Section~\ref{sec:data}, we define the \textit{label of an utterance} as \textit{the list of labels of all sentences in it}. Using this definition of \textit{label}, we report prevalence-weighted macro-average of F1 scores for DAs / emotions (\textbf{\textit{DAs / Emotions-F1}}). 

Table~\ref{tab:semantic-level-automatic-results} shows the results of semantic-level automatic evaluation. The vanilla models have higher DAs-F1 and Emotions-F1 than the \textit{min length} models. It shows that by enforcing the models to generate long responses, we get sacrificed semantic-level agreement with human references. On the other hand, our SE-GPT-2 and CDialGPT-2 models largely improve the semantic-level metrics. This result shows that explicitly planning the semantic variables helps improve the semantic-level performance when generating long responses. 

While the GPT-2 models generally outperform the CDialGPT-2 models in terms of the conventional metrics in Table~\ref{tab:word-level-automatic-results}, their semantic-level performance is comparable. It indicates that pretrained weights have a larger influence at the token level than at the semantic level. Thus, it is probable that one needs to consider mechanisms besides pretraining (e.g., knowledge about the semantic transitions) to improve the semantic-level performance.

\subsection{Controllability}
\label{subsec:controllability}
A desirable feature of our method is its controllability, e.g., the semantic plans can be \textit{intervened} by rules and humans. To evaluate controllability, we compute the similarity between model predictions and human references when providing the semantic plans in the original test data. Higher similarity indicates better controllability. Table~\ref{tab:word-level-automatic-results} shows that the token-based and embedding-based metrics are greatly improved when semantic plans are provided. It shows that the semantic variables can effectively guide the response generation. Table~\ref{tab:word-level-automatic-results} also shows that the controlled model has improved diversity over the non-controlled models. This observation shows that, in terms of diversity, there is still space for the \textit{planning} module $P_{\theta}$ to be improved. Table~\ref{tab:semantic-level-automatic-results} shows that more than 90\% topical words are generated in the guided responses, while the DA and emotion F1 scores are around 90\% and 70\%. In real-world scenarios, such controllability allows us to guide the chatbot when displaying certain contents, DAs, or emotions. 

\begin{table*}[t]
\centering
\small
\begin{tabular}{@{}l@{}ccccccc@{}}
\toprule
~  & \multicolumn{4}{c}{Non-semantic metrics} & \multicolumn{3}{c}{Semantic-level metrics} \\
\cmidrule(r){2-5} \cmidrule(l){6-8}
~  			                                & BLEU-2    & BLEU-3    & Emb-Avg   & Dist-2 \% & Topical-R & DAs-F1             & EMOs-F1  \\
\midrule
SE-GPT-2 (ours)                            & \bf 6.2   & \bf 3.1   & \bf 0.806 & \bf 11.57 & \bf 13.1  & 52.1  & 22.2 \\ 
\quad w/o \textit{understanding}            & 6.1       & \bf 3.1   & 0.805     & 11.27     & 12.7      & 52.2  & \bf 22.8 \\ 
\quad w/o \textit{planning}                 & \underline{5.3}   & \underline{2.7}       & 0.805 & \underline{9.01} & 5.0      & \underline{46.1}       & \underline{19.4} \\
\quad w/o \textit{repetition constraint}    & 6.1       & \bf 3.1   & 0.804     & 11.46     & 12.9      & 52.0      & 21.8 \\
\quad w/o \textit{topical words min length} & \underline{5.3}       & 2.8       & \underline{0.794}     & 9.51      & \underline{2.7}       & 52.0  & 21.3 \\
\midrule
SE-CDialGPT-2 (ours)                       & 6.1       & \bf 3.1   & \bf 0.806 & \bf 10.53 & \bf 12.7  & 52.1  & \bf 22.4 \\
\quad w/o \textit{understanding}            & \bf 6.2       & \bf 3.2   & \bf 0.806 & 10.41 & 12.1      & 52.1      & 22.3 \\
\quad w/o \textit{planning}                 & 5.2       & \underline{2.6}       & \bf 0.806 & \underline{8.02} & 4.4      & \underline{45.6}      & \underline{19.1} \\
\quad w/o \textit{repetition constraint}    & 5.7       & 2.9       & \underline{0.794}     & 10.17     & 11.8      & 52.1      & 22.2 \\
\quad w/o \textit{topical words min length} & \underline{5.1}       & 2.7       & \underline{0.794}     & 8.52      & \underline{2.5}       & 52.1      & 21.6 \\
\bottomrule
\end{tabular}
\caption{\label{tab:ablation-results} Ablation studies. Best results are shown in \textbf{bold}, and worst results are \underline{underlined}.}
\end{table*}

\subsection{Human Evaluation}
\label{subsec:human-evaluation}

We also perform human evaluations to validate our method. We recruit annotators from a third-party crowd-sourcing platform to conduct the human evaluation. Each model prediction is scored based on its 1) relevance and coherence to the dialogue history, 2) informativeness, and 3) engagingness. We use 400 random test samples, and each model output is scored independently by three annotators using a 3-point Likert scale. Since the GPT-2 backbone generally has better automatic results, we only compare models that use GPT-2 as the backbone. As the upper bound, we also include the gold response as a baseline (note that we tell annotators that all responses are produced by machines).

Human evaluation results are presented in Table~\ref{tab:human-results}. The free-marginal kappa score for relevance and coherence, informativeness, and engagingness is 0.79, 0.70, and 0.75, respectively, which all show substantial agreement. Human evaluation generally echoes the observations in automatic evaluation. Specifically, the \textit{min length} model outperforms the vanilla model, and our method outperforms the baselines in terms of all aspects. The improved relevance and coherence may result from the fact that understanding and planning help strengthen the semantic-level transition that flows throughout the conversation. The improved informativeness shows that the semantic plans with the constraints can mitigate the problem of generic responses \cite{LiGBGD16}. We also note that the engagingness is \textit{not significantly} improved. It can be attributed to the fact that the engagingness is related to the \textit{way of presentation} and is less strongly related to the semantic-level property of conversation. \textit{Unlike} prior works that increase the \textit{model sizes} and \textit{amounts of pretraining data}, our method uses the \textit{same} model size and amount of data as the baselines. Thus, the significant improvements of relevance, coherence, and informativeness come solely from the modular framework. 

\subsection{Model Analysis}
\label{subsec:model-analysis}

We conduct ablation studies to understand the contribution of some components: 1) w/o \textit{understanding} and 2) w/o \textit{planning} remove \textit{Human}'s and \textit{Machine}'s semantic variables from Eq.~(\ref{eq:order-of-sequences}), respectively. 3) w/o \textit{topical words min length} and 4) w/o \textit{repetition constraint} remove the minimum length and the repetition constraint for the planning of topical words, respectively. Results are presented in Table~\ref{tab:ablation-results}. After removing the \textit{understanding} component, we observe slightly dropped diversity and topical word recall, which suggests that explicitly tracking the topical words of the dialogue history improves the diversity and topical-level performance. Removing the \textit{planning} component leads to a large drop in terms of nearly all metrics, which shows that \textit{planning} contributes the most to our method. Although the repetition constraint does not significantly influence the performance, we view it as necessary since it avoids the repetition problem observed in preliminary experiments. The w/o \textit{topical words min length} ablations have largely dropped topical words recall, which shows that the minimum length constraint for topic planning leads to more informative topical words to be generated.

\begin{table}[t]
\centering
\small
\begin{tabular}{@{}l@{}ccc@{}}
\toprule
& Topical-F1 & DAs-F1 & EMOs-F1 \\
\midrule
SE-GPT-2         & 98.7  & 93.0  & 77.6  \\
SE-CDialGPT-2    & 98.5  & 92.7  & 76.4  \\
\bottomrule
\end{tabular}
\caption{\label{tab:automatic-results-planning} Evaluation for the \textit{understanding} module}
\end{table}

We also evaluate the performance of the \textit{understanding} module to investigate whether it can provide reliable semantic-level summaries of the dialogue history. We compute the \textit{\textbf{DAs / Emotions-F1}} based on the \textit{understanding} module's outputs and the semantic variables in X-\textsc{Weibo}. For each sample, we compute the F1 score of predicted topical words and topical words in X-\textsc{Weibo}, and we average them over all test samples, denoted as \textit{\textbf{Topical-F1}}. Table~\ref{tab:automatic-results-planning} shows that the \textit{understanding} module achieves high performances for topical words and DAs. The comparably lower emotions F1 can be attributed to the fact that emotions are more ambiguous than DAs, which is analyzed in Section~\ref{sec:data}. The overall performance shows that the \textit{understanding} module can provide reliable semantic-level summaries of the dialogue history. 

\section{Discussion}
\label{sec:discussion}

As discussed in Section~\ref{sec:data}, human annotation of the semantic variables faces the \textit{scalability} problem when applied to large-scale conversation data (e.g., over 3.3M utterances in our dataset). To address this challenge, we provide a preliminary attempt by using automatic annotation with a careful selection of training data. Although we have provided a human verification of our annotation, such annotation unavoidably introduces the \textit{distributional shift} between the classifier's training data and the conversation data. Future researches may explore \textit{semi-supervised learning} methods, e.g., variational inference \cite{KingmaW13} and domain adaptation methods \cite{Ramponi20Survey}, which helps address the distributional shift problem between datasets.

Our experimental results show that modeling \textit{understanding} and \textit{planning} with semantic variables improves open-domain dialogue modeling. Promising results suggest that \textit{understanding} and \textit{planning}, which are less investigated for open-domain dialogues, should get more attention. Specifically, the following directions can be investigated. 
\begin{itemize}
    \item Improved understanding and planning variables with end-to-end training. BLEU scores and PPL in Table~\ref{tab:word-level-automatic-results} (w/ \textit{gold variables}) show that there is a large space left for the response to be further annotated. The results suggest finer-grained variables to be studied for open-domain dialogue modeling. 
    \item Controllable dialogue generation. Since \textit{planning} disentangles semantic-level and token-level variations, it is natural to apply it to controllable dialogue generation, e.g., a speaker's \textit{style}, \textit{stance}, and \textit{bias} could be disentangled into the semantic level and the token level. 
\end{itemize}

We adopt the DA scheme used in the DailyDialog dataset \cite{LiSSLCN17}, while it should be noted that a wide range of fine-grained DAs have been proposed \cite{MezzaCSTR18,BuntPGPFKP20}. Besides semantic meaning, human conversations also convey pragmatic meaning and implications that require commonsense reasoning \cite{BosselutRSMCC19,SapRCBC19}. These messages are also beyond the token level and even go beyond the semantic level. Promising results in our experiments may encourage future research to explore such higher-level explainability, e.g., explicit \textit{understanding} and \textit{planning} aided by pragmatic and commonsense reasoning. 

\section{Conclusions}
In this paper, we propose \textit{semantic-enhanced finetuning} to bridge the pretrained language models with the modular dialogue modeling paradigm for \textit{open-domain} dialogue systems. Our method leverages the response generation ability of pretrained models while being explainable and controllable. To address the scalability issue of semantic annotation, we present X-\textsc{Weibo}, which is automatically annotated with topical words, DAs, and emotions for each utterance. Experimental results show that our method outperforms strong baselines in terms of automatic and human evaluations and has considerable controllability. Finally, we discuss ways to further address the scalability problem of semantic annotation and possible future works to consider fine-grained semantic variables, controllability, and pragmatic and commonsense reasoning. 

\section*{Acknowledgments}
This work was partly supported by the NSFC projects (Key project with No. 61936010 and regular project with No. 61876096). This work was also supported by the Guoqiang Institute of Tsinghua University, with Grant No. 2019GQG1 and 2020GQG0005. 

 
\bibliographystyle{IEEEtran}
\bibliography{tacl2018}


\section{Biography Section}
%
%

\begin{IEEEbiography}[{\includegraphics[width=1in,height=1.25in,clip,keepaspectratio]{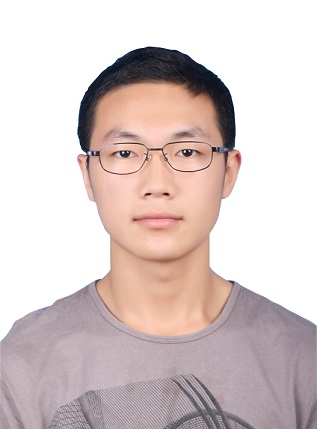}}]{Yinhe Zheng}
received his Ph.D. degree from China University of Geosciences (Beijing), Beijing, China, in 2017. He worked as a Post Doctor in a joint program of the Department of Computer Science and Technology, Tsinghua University, and Samsung Research China - Beijing (SRCB). His research interests include natural language processing and dialogue system, especially the tasks related to natural language understanding and natural language generation. He is the recipient of the Wuwenjun AI Award in 2019.
\end{IEEEbiography}

\begin{IEEEbiography}[{\includegraphics[width=1in,height=1.25in,clip,keepaspectratio]{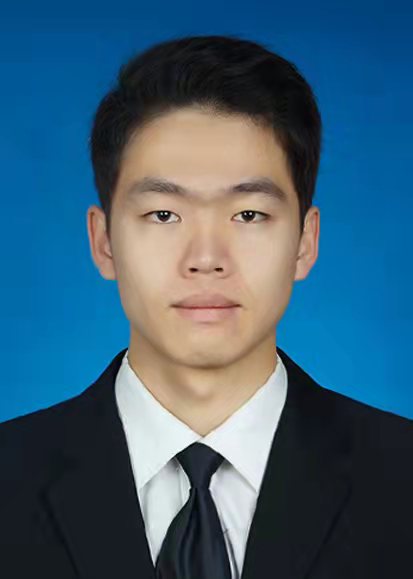}}]{Yida Wang} received his MS degree from Tsinghua University, Beijing, China in 2021. He is currently working as an investment researcher in an asset management company. His research interests include natural language processing and quantitative investment, especially the task realated to dialogue system and portfolio optimization. 
\end{IEEEbiography}

\begin{IEEEbiography}[{\includegraphics[width=1in,height=1.25in,clip,keepaspectratio]{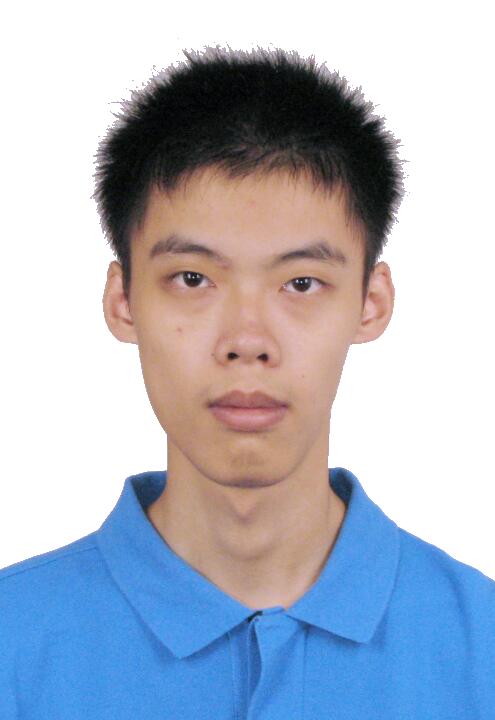}}]{Pei Ke} received the B.E. degree from Zhejiang University, China, in 2017. He is a PhD student at the Department of Computer Science and Technology, Tsinghua University. His research interests include natural language generation, dialogue systems, and sentiment analysis.
\end{IEEEbiography}

\begin{IEEEbiography}[{\includegraphics[width=1in,height=1.25in,clip,keepaspectratio]{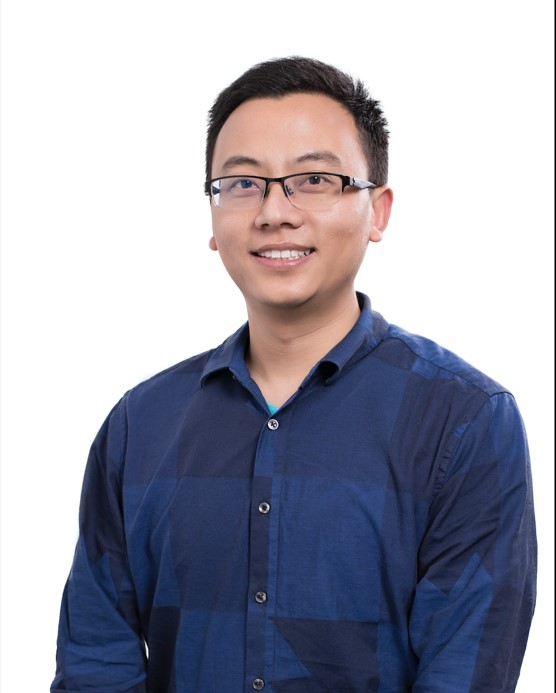}}]{Zhenyu Yang} received his B.Eng. and Ph.D. degrees from the University of Science and Technology of China, Hefei, Anhui, China, in 2005 and 2010, both in computer science. He is currently the Chief Research Scientist with the Center of Xiaobu Assistant, Guangdong OPPO Mobile Telecommunications Corp., Ltd, Shenzhen, China. His research interests include dialogue systems, metaheuristics and various real-world applications.
\end{IEEEbiography}

\begin{IEEEbiography}[{\includegraphics[width=1in,height=1.25in,clip,keepaspectratio]{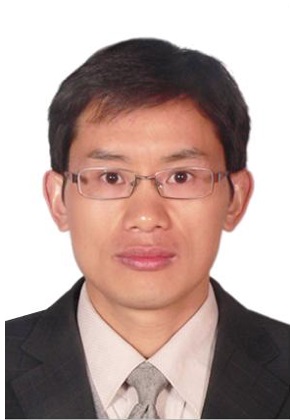}}]{Minlie Huang}
received his Ph.D. degree from Tsinghua University, Beijing, China, in 2006. He is currently an Associate Professor with the Department of Computer Science and Technology, Tsinghua University. His research interests include natural language processing, particularly in dialog systems, reading comprehension, and sentiment analysis. He has published more than 60 papers in premier conferences and journals (ACL, EMNLP, AAAI, IJCAI, WWW, SIGIR, etc.). His work on emotional chatting machines was reported by MIT Technology Review, the Guardian, Nvidia, and many other mass media. He serves as standing reviewer for TACL, area chairs for ACL 2020/2016, EMNLP 2019/2014/2011, and Senior PC members for AAAI 2017-2020 and IJCAI 2017-2020, and reviewers for TASLP, TKDE, TOIS, TPAMI, etc. He is a nominee of ACL 2019 best demo papers, the recipient of IJCAI 2018 distinguished paper award, CCL 2018 best demo award, NLPCC 2015 best paper award, Hanvon Youngth Innovation Award in 2018, and Wuwenjun AI Award in 2019. He was supported by a NSFC key project , several NSFC regular projects, and many IT companies.
\end{IEEEbiography}

%
%

\vfill

\end{document}